%% file: root.tex
\documentclass[letterpaper, 10 pt, conference]{ieeeconf}  
\IEEEoverridecommandlockouts                              
\usepackage{booktabs}
\usepackage{multirow}
\usepackage{graphicx}
\usepackage{lipsum}
\usepackage{xcolor}
\usepackage{cite}
\graphicspath{ {./images/} }
\usepackage{caption}
\usepackage{pgffor}
\usepackage{xcolor}   % For colors
\usepackage{pifont}   % For check marks and crosses
\definecolor{darkgreen}{rgb}{0.0, 0.6, 0.0}  % Darker green (adjust values as needed)
\usepackage{amssymb}

\usepackage{graphics} % for pdf, bitmapped graphics files
\usepackage{epsfig} % for postscript graphics files
\usepackage{times} % assumes new font selection scheme installed
\usepackage{amsmath} % assumes amsmath package installed
\usepackage{amssymb}  % assumes amsmath package installed
\usepackage{hyperref}
\usepackage{subcaption}
\usepackage{algorithm}
\usepackage{array}
\usepackage{booktabs}

\title{\LARGE \bf
DexHub and DART: Towards Internet Scale Robot Data Collection
}

\author{Younghyo Park$^{1}$, Jagdeep Singh Bhatia$^{1}$, Lars Ankile$^{1}$, and Pulkit Agrawal$^{1}$
\thanks{$^{1}$Improbable AI Lab at MIT
        {\tt\small younghyo@mit.edu}}%
}

\newif\ifShowComments
\ShowCommentstrue

\newcommand{\platform}{DexHub}
\newcommand{\interface}{DART}

\usepackage{graphicx}
\usepackage{etoolbox}
\newcommand{\insertfig}{\vspace{0.5em}\includegraphics[width=\linewidth]{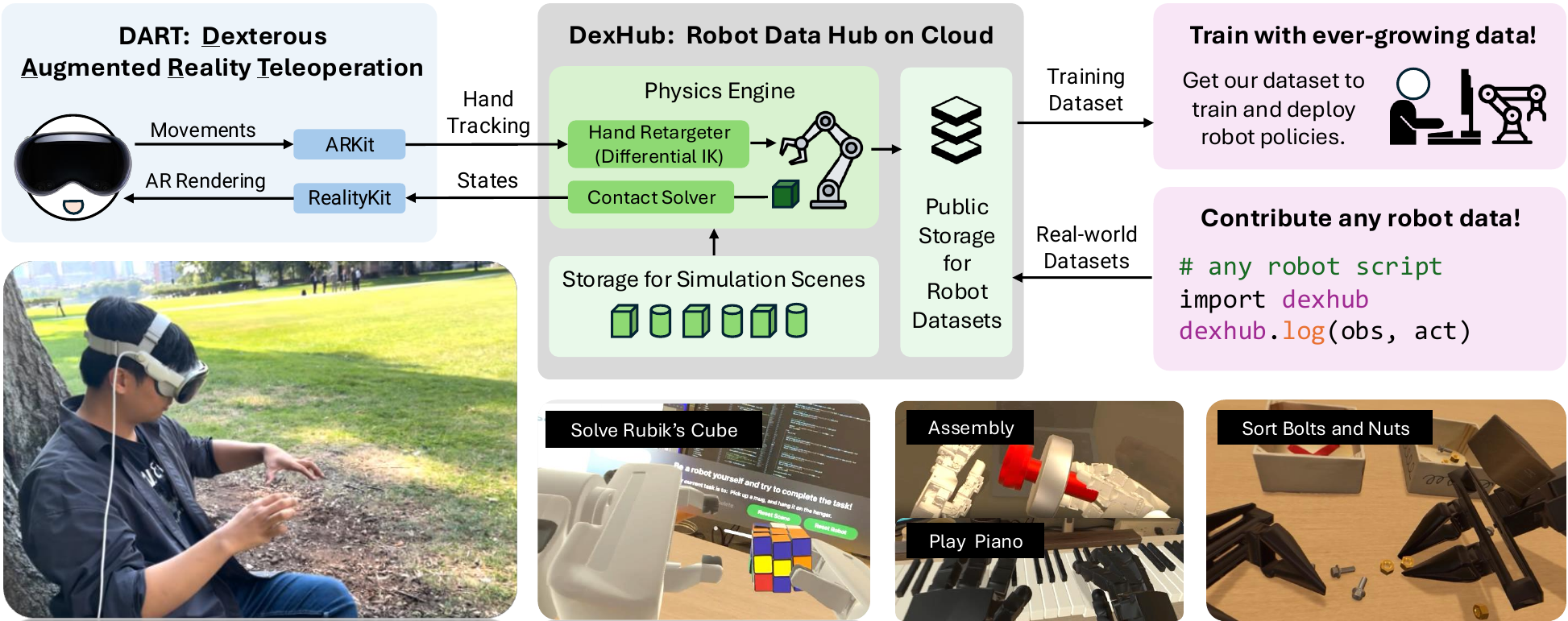}\captionof{figure}{We present \textbf{\interface{}}, \underline{D}exterous \underline{A}ugmented \underline{R}eality \underline{T}eleoperation system, enabling intuitive, low-latency teleoperation with cloud-hosted simulation. Through a user study, we found that \interface{} enables first-time robot teleoperators to achieve 2.1$\times$ faster data collection throughput with significantly lower physical fatigue than existing real-world teleoperation platforms. To further support scaling up data collection efforts in the community, we also release \textbf{\platform{}}, a cloud-hosted data hub for robot learning where data collected in \interface{} is automatically stored. 
\url{https://dexhub.ai/project}}\label{fig:title_figure}
\vspace{-7pt}
}
\usepackage{caption} 
\makeatletter
\apptocmd{\@maketitle}{\centering\insertfig}{}{}% insert the figure after authors
\makeatother

\begin{document}

\maketitle

\thispagestyle{empty}
\pagestyle{empty}

\begin{abstract}
The quest to build a generalist robotic system is impeded by the 
scarcity of diverse and high-quality data. 
While real-world data collection effort exist, requirements for robot hardware, physical environment setups, and frequent resets significantly impede the scalability needed for modern learning frameworks.
We introduce \interface{}, a teleoperation platform designed for crowdsourcing that reimagines robotic data collection by leveraging cloud-based simulation and augmented reality (AR) to address many limitations of prior data collection efforts. Our user studies highlight that \interface{} enables higher data collection throughput and lower physical fatigue compared to real-world teleoperation. We also demonstrate that policies trained using \interface{}-collected datasets successfully transfer to reality and are robust to unseen visual disturbances. All data collected through \interface{} is automatically stored in our cloud-hosted database, \platform{}, which will be made publicly available upon curation, paving the path for \platform{} to become an ever-growing data hub for robot learning.
\url{https://dexhub.ai/project}
\end{abstract}

\input{intro}
\input{related-works}

\input{arteleop}

\input{experiments}

\input{dexhub}
\input{conclusion}

\section*{ACKNOWLEDGMENT}

We thank the members of the Improbable AI lab for the helpful discussions and feedback on the paper. We are grateful to MIT Supercloud and the Lincoln Laboratory Supercomputing Center for providing HPC resources. This research was partly supported by Hyundai Motor Company, DARPA Machine Common Sense Program, the MIT-IBM Watson AI Lab.
DARPA Machine Common Sense Program, the MIT-IBM Watson AI Lab, and the National Science Foundation under Cooperative Agreement PHY-2019786 (The NSF AI Institute for Artificial Intelligence and Fundamental Interactions, http://iaifi.org/). We acknowledge support from ONR MURI under grant number N00014-22-1-2740. Research was sponsored by the Army Research Office and was accomplished under Grant Number W911NF-21-1-0328. The views and conclusions contained in this document are those of the authors and should not be interpreted as representing the official policies, either expressed or implied, of the Army Research Office or the U.S. Government. The U.S. Government is authorized to reproduce and distribute reprints for Government purposes notwithstanding any copyright notation herein.

\newpage
\bibliographystyle{IEEEtran}
\bibliography{IEEEabrv,references}

\end{document}

%% file: intro.tex
\section{Introduction}
Robotics has seen impressive progress with the advent of learning-based control. However, a major bottleneck is the lack of diverse and high-quality data for training robust and generalizable robot policies. Access to an internet-scale robotics dataset that continually and rapidly grows with data coming from everywhere in the world will be ideal --- just like how people easily upload language, images, and videos on the internet. Despite recent efforts~\cite{o2024open, fang2024rh20t, ebert2021bridge, khazatsky2024droid}, we are not there yet. In this paper, we examine and address many key bottlenecks in achieving this dream. 

Consider collecting data to perform a given task, such as moving dishes from the sink to the dishwasher. The data collector's first challenge is \textbf{\textit{setting up the environment}}. There are two options: physically construct a kitchen in the lab around the robot or physically move the robot to an actual kitchen. Neither is easy to scale as data will be needed from many kitchens.

Once the environment is set up, operating the robot leads to the second challenge -- \textbf{\textit{observing and understanding}} the scene. 
For instance, due to visual occlusions and lack of tactile feedback, operators may be unable to sense an object's motion resulting from the robot's action. Further, if the teleoperation is remote, it adds additional challenges originating from network delays, limited field of view, and visual artifacts. Such challenges can slow down operators and often prevent them from performing dynamic or precise tasks.

If the data collector succeeds at resolving the first two challenges and moves all the dishes from the sink to the dishwasher to complete the exemplar task, a third obstacle emerges: all the dishes must be returned to the sink to collect a new trajectory! In addition to being time-consuming, this \textbf{\textit{resetting}} is physically and mentally exhausting as operators must context-switch between robot control and environment setup. Ensuring that each reset presents the robot with a diverse range of scenarios is also mentally taxing requiring imagination. 

What makes the experience even worse for the operators is the need to \textbf{\textit{repeat}} the process of \textit{teleoperating} and \textit{resetting} a large number of times. 
The number of required demonstrations and the need for diversity in demonstrations scales with the task complexity and the extent of required generalization. 
Unfortunately, humans are known to lose focus when performing a repetitive job \cite{hausser2014experimental}. 

Finally, say the operator has finished collecting a few hundred demonstrations. How does the recorded data get \textbf{\textit{processed, stored, and used}}? It is common to store collected demonstrations on a local machine or a \textit{private} cloud, which is often not shared to general publicly unless someone explicitly requests it. 

\vspace{1em}

These issues, all combined, make existing data collection methods struggle to scale up without operator fatigue. Making everything worse, the data collected in real-world has limited applicability in terms of policy training methods; reinforcement learning, for instance, cannot be easily applied on top of real-world demonstrations as it lacks a digital twin where virtual agents can freely explore and self improve its performance. A data collection method that (a) can be easily parallelized and crowd-sourced with minimal hardware requirements and (b) wide range of policy training pipelines can be applied can get closer to the needed scale and diversity of robot data. 

To that end, we introduce \textbf{\interface{}}, a \underline{D}exterous \underline{A}ugmented \underline{R}eality \underline{T}eleoperation system, enabling \textit{anyone} in the world to teleoperate robots in simulation with an intuitive, game-like AR interface. Connected to a cloud-hosted simulation, \interface{} allows users to collect demonstrations for an unlimited number of scenes in one sitting without having to physically set up environments or physically move robots to different places. \interface{}'s high-fidelity AR rendering allows users to observe the scene in great detail with minimal occlusion, enabling teleoperation of complex tasks. \interface{} also allows users to reset the environment with a click of a button, removing the taxing process of physically resetting the scene. 

As a result, our user study shows that \interface{} achieves 2.1$\times$ \textbf{faster data collection throughput} with significantly less physical and cognitive fatigue on tasks requiring fine-grained control compared to most existing robot data collection pipelines. Our  experiments also highlight the unmatched benefits of collecting demonstrations in simulation over the real world. Simulation-trained policies achieve higher robustness than real-world trained policies due to data augmentation and randomization strategies only possible in simulation. Finally, all robot demonstrations collected through \interface{} are automatically stored and logged to our public cloud-hosted database, \textbf{\platform{}}, which serves as an open-sourced data hub for robot learning. 

Our key contributions are outlined as follows: 

\begin{enumerate} 
\item In Sec. \ref{sec:art}, we introduce \interface{}, a novel AR-based teleoperation platform, and detail its system architecture and supported features. We also showcase the diversity of tasks we can perform with \interface{}, unlocked by enhanced teleoperation experience. 
\item In Sec. \ref{sec:user-study}, we analyze the impact of different teleoperation interface design choices through user study. We show that \interface{} enables higher data collection throughput and lower fatigue than alternatives.
\item In Sec. \ref{sec:sim2real}, we show that policies trained with data collected via \interface{} can be effectively transferred to the real world and are more robust than those trained with real-world demos. 
\item In Sec. \ref{sec:dexhub}, we provide an brief overview of the proposed \platform{} platform that serves as a central hub for hosting large-scale robot demonstrations generated by \interface{}.
\end{enumerate}

% \vspace{-1em}

\vspace{5pt}

%% file: related-works.tex
\section{Related Works}

\subsection{Large-Scale Robot Data Collection Efforts}

Addressing the need for large-scale datasets in robotics, there have been two primary approaches within the community. The first approach, as exemplified by projects like \cite{o2024open}, focuses on gathering existing datasets from various robotics institutes worldwide into a single place. These initiatives involve a central team overseeing the data gathering, post-processing, and release. The second approach involves teams actively collecting large-scale datasets themselves by teleoperating robots in real-world environments. For example, \cite{fang2024rh20t} collected 110k trajectories for diverse tasks through real-world teleoperation with the help of volunteer participants. Similarly, \cite{ebert2021bridge} created a dataset of 60k trajectories using a low-cost robotic arm. Most recently, \cite{khazatsky2024droid} have released 76k demonstrations across 564 scenes using a Franka Panda attached to a mobile platform. These efforts all unanimously highlight the value of large datasets in improving the performance of trained policies. 

However, we argue that relying on disconnected, project-level efforts to create such datasets is not a scalable solution for the robotics community. The episodic, labor-intensive collection efforts seen in these examples fail to mirror the organic growth of language and vision datasets on the internet. Furthermore, these datasets are limited in scope, primarily focused on single-arm robots with parallel jaw grippers, neglecting the richness of bimanual or dexterous manipulations. Finally, these datasets are collected exclusively in real-world settings, overlooking the significant potential of simulation as a data source. Simulation allows for the refinement and augmentation of human-collected -- and therefore possibly suboptimal -- datasets through online reinforcement learning using massively parallelizable simulation environments \cite{makoviychuk2021isaac}. Such refinement can address the potential performance saturation often observed on policies trained only with supervised learning \cite{pmlr-v9-ross10a, ross2011reduction, zhaoaloha, ankile2024imitation}.

\subsection{Collecting Robotic Dataset in Simulation}

Using simulation as an alternative environment for collecting demonstrations has been explored in the community. For example, \cite{qin2023anyteleop} utilized webcams attached to laptops to allow users to teleoperate various robot morphologies in simulation. \cite{mosbach2022accelerating} employed a VR interface where humans control simulated dexterous hands, while specialized exoskeletons capture their hand movements. More recently, with advancements in VR devices, \cite{iyer2024open, cheng2024open} have demonstrated similar technical stacks that no longer require external hand trackers, but instead utilize the built-in capabilities of modern VR/AR devices to capture hand movements. All aforementioned systems use stereo rendering streams as a source of visual feedback.
However, relying on raw visual streams of simulated renderings inevitably creates a noticeable latency in network communication, forcing designers to trade-off visual fidelity and latency to maintain real-time performance. 
The use of Augmented Reality (AR) objects instead as a visual scene representation, on the other hand, has not yet been thoroughly explored as a solution to this problem. 
Finally, no existing platform has fully leveraged simulation's potential by making data collection widely accessible and available to the general public -- particularly to those without specialized knowledge in robotics or the ability to set up simulation servers.

\subsection{Augmented Reality for Robot Data Collection}

Augmented Reality (AR) has been explored as a valuable tool to support the data collection process for robots. For instance, \cite{duan2023ar2} leveraged mobile device AR capabilities to develop a waypoint-based teaching pendant using a virtual robot. Similarly, \cite{chen2024arcap} used AR renderings to provide visual cues of robot behaviors while recording human motions in the real world. \cite{van2024puppeteer} also employed AR-rendered robots to guide the teleoperation process for real-world robots. However, none of these works fully leveraged AR’s potential to teleoperate virtual robots in simulation through a tightly integrated control-sensory feedback loop, particularly with an emphasis on large-scale, crowd-sourced data collection.

%% file: arteleop.tex
\section{\interface{}: Teleoperating Robots in Sim via AR}
\label{sec:art}

This section details the system architecture of \interface{} and its benefits (Sec \ref{sec:arteleop-sys-arch}). We then introduce the main features of the platform (Sec \ref{sec:arteleop-sys-feat}), which are designed to maximize the platform's capability (Sec \ref{sec:capability}) and enhance user experience. 

\subsection{System Architecture} \label{sec:arteleop-sys-arch}

\interface{}'s key components facilitate intuitive, low-latency teleoperation available for anyone worldwide. 

\subsubsection{Simulation Assets as AR Objects} 

Enabled by Apple's RealityKit, \interface{} presents all assets in simulation environments, including robots, as photo-realistic AR objects overlayed over each operator's real-world environment. Handling visualization locally on the AR device (a) removes unnecessary latency from transmitting large image data packets and (b) significantly improves the real-timeliness of the simulation by removing the compute-intensive rendering layer. Variation in latency critically impacts the user's data collection throughput and cognitive fatigue, as highlighted by our user study (See Sec. \ref{sec:user-study}). 

\subsubsection{Low-Latency Communication}

Communication between the AR device, i.e., Apple Vision Pro, and the cloud-hosted simulation is handled via gRPC, which facilitates low-latency, \textit{asynchronous} bidirectional data transfer. The AR device sends hand-tracking data to the simulation, and asynchronously receives the simulation state. Table \ref{table:packet-sizes} highlights the reduced network load of our approach compared to a typical setting where real-world or simulated camera streams are transmitted over the network. Even in the most adversarial case, where robots have $n=58$ joints and simulation scenes contain $m=50$ objects, the data packet size is over \textbf{1,000$\times$ smaller} than that required for existing teleoperation frameworks.

\subsubsection{Cloud-Hosted Simulation}

The robot simulation is powered by MuJoCo~\cite{todorov2012mujoco} and dynamically launched on AWS Elastic Container Registry (ECR) as users join. Each simulation instance runs in the cloud, enabling open access and low user setup costs. Due to compact packet sizes (Table \ref{table:packet-sizes}), cloud-hosting does not critically impact the overall latency of our platform compared to local-hosting, as evidenced in Table \ref{table:sys-eval}.  

\begin{table}[t]
\centering
\begin{tabular}{@{}ll|c|c@{}}
\toprule
 &  & \interface{} (Ours) & \begin{tabular}[c]{@{}c@{}}Others\end{tabular} \\ \midrule
\multirow{3}{*}{\begin{tabular}[c]{@{}l@{}}Human \\ $\rightarrow$ Robot\end{tabular}} & \multirow{2}{*}{Data Type} & Hand Tracking & \begin{tabular}[c]{@{}c@{}}Hand and Head\\Tracking \end{tabular}\\ \cmidrule(l){3-4} 
 &  & \begin{tabular}[c]{@{}c@{}}25 Hand Keypoints\\ $\times$ SE(3) \end{tabular} & \begin{tabular}[c]{@{}c@{}}(25 Hand Keypoints\\ + 1 Head) $\times$ SE(3) \end{tabular} \\ \cmidrule(l){2-4} 
 & Packet Size & 0.7kB & 0.728kB \\ \midrule
\multirow{3}{*}{\begin{tabular}[c]{@{}l@{}}Robot \\ $\rightarrow$ Human\end{tabular}} & \multirow{2}{*}{Data Type} & Oracle Sim States & Stereo RGB image \\ \cmidrule(l){3-4} 
 &  & \begin{tabular}[c]{@{}c@{}}$n$ joints $\times$ $\texttt{float}$\\ $m$ objects $\times$ SE(3) \end{tabular} & \begin{tabular}[c]{@{}c@{}}2$\times$(480$\times$640$\times$3)\\ $\texttt{uint8}$\end{tabular}\\ \cmidrule(l){2-4} 
 & Packet Size & \textbf{1.6kB} & \textbf{1843.2kB} \\ \bottomrule
\end{tabular}
\caption{We highlight \interface{}'s \textbf{1,000$\times$ reduction} in network packet size between robot and operator's AR device compared to existing frameworks. $n=58$, $m=50$ assumed for \interface{}. }
\label{table:packet-sizes}
\vspace{-1em}
\end{table}

\subsubsection{Hand Tracking and Mapping} 

\begin{figure}[t]
\centering
\includegraphics[width=\columnwidth]{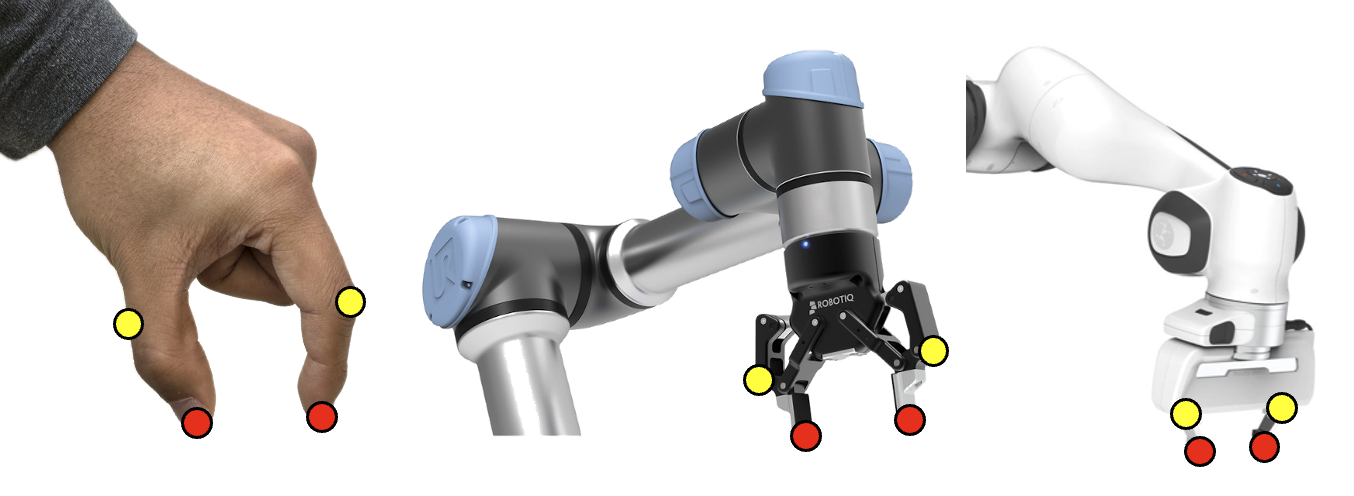}
\caption{4 finger keypoints used as tracking points for robots with parallel-jaw grippers.}
\label{fig:finger-mapping}
\end{figure}

\interface{} leverages Apple's ARKit to track poses of hand and wrist keypoints. We use a subset of detected keypoints, which fully determine the end-effector and finger movements, as target points for robots to track. Specifically, for robot systems with parallel-jaw grippers, we use the $xyz$ position of 4 finger key points 
as tracking targets, which fully determine the SE(3) pose of the robot's end-effector (Fig. \ref{fig:finger-mapping}). \interface{} uses differential inverse kinematics \cite{pink2024} by defining position-only tracking costs for each keypoints, $e(\mathbf{p})$. We additionally apply basic  safety constraints, i.e., self-collision avoidance, expressed as $d(q)$. The resulting optimization problem is as follows,
\begin{equation}
    \begin{aligned} 
        \min_{v\in c} & \sum_{\mathbf{p}\in\mathcal{P}} \| J_e(q)v + \alpha e(\mathbf{p})\|^2 \\
        \text{s.t.}~ & v_{\text{min}}(q) \leq v \leq v_{\text{max}}(q), ~d(q) > 0.
    \end{aligned}
\end{equation}
For dexterous five-fingered hands, we use six position-only keypoints -- five from the fingertips and one from the wrist.

\begin{table}[]
\centering
\begin{tabular}{@{}ccc@{}}
\toprule
                         & \begin{tabular}[c]{@{}c@{}}Cloud-Hosted\\ Simulation\end{tabular} & \begin{tabular}[c]{@{}c@{}}Local Machine\\ and Local Network\end{tabular} \\ \midrule
CPU                      & AWS EC2 C7i                                                       & i9-13900k                                                                 \\ \midrule
Packet Travel Time  & 15.4 ms                                                           & 10.3 ms                                                                   \\
Simulation Step          & 1.8 ms                                                            & 1.6 ms                                                                    \\ \midrule
Total                    & 17.2 ms                                                           & 11.9 ms                                                                   \\ \bottomrule
\end{tabular}
\caption{Comparing the time profile of our system running on the cloud v/s hosted on a local machine. AWS instance was hosted on us-east-1, connected from Boston. 
}
\label{table:sys-eval}
\vspace{-1.5em}
\end{table}

\subsection{System Features} \label{sec:arteleop-sys-feat}

\interface{} supports a wide range of features to enhance the teleoperation experience while maintaining low setup costs, allowing \textit{anyone} to participate in robotics data collection. Although it is currently developed for Apple Vision Pro, Apple's AR device, the design decisions can also be developed and applied for lower cost AR devices. 

\subsubsection{Pre-Designed Robots and Scenes} 
 Out-of-the-box, \interface{} supports many robots: multiple end-effectors (Robotiq 2F-85 gripper, Panda Hand, Allegro Hand) can be attached to bimanual setup of Franka Research 3 or UR-5. Unitree Humanoid Series (G1) and ALOHA \cite{zhao2023learning} are also supported. High-fidelity MuJoCo models of these robots were provided by \cite{menagerie2022github}. 

\subsubsection{Importing Custom Scenes} 
Users can import custom simulation environments and assets to extend the platform's capabilities further. Assets can be uploaded through our online portal (\url{https://dexhub.ai/}) and accessed via \interface{} App on VisionOS App Store.

\subsubsection{One-Click Reset} 

\interface{} includes an efficient task-resetting feature in simulation. Users can reset the environment with a single click of a button, significantly reducing operator fatigue and increasing  data collection throughput.

\subsubsection{Instant Task Switching}

In addition to resetting a single scene, \interface{} enables quick switching between various tasks and simulation environments. This functionality minimizes the operator’s mental fatigue that arises from repetitively performing the same task, allowing for a more engaging data collection experience.

\subsection{Capability and Task Diversity}
\label{sec:capability}

\interface{} is capable and versatile. It supports a wide range of tasks, from simple object manipulation to complex, precise, and dexterous maneuvers, as highlighted in Figure \ref{fig:title_figure}. These examples and those below illustrate the platform's potential to support various research and practical applications in robotics.

\begin{itemize}
\item Fine motor skills: e.g., picking up small objects.
\item Household chores: e.g., hanging mugs on a rack. 
\item Dexterous Manipulation: e.g., solving a Rubik's cube.
\end{itemize}

%% file: experiments.tex
\section{Experiments}
\label{sec:experiments}

Our experiments address two key questions: 

\begin{enumerate}

\item \textit{How \textbf{intuitive} is \interface{} for robotics novices to use?} We conduct a formal user study to assess the platform's accessibility to individuals without robotics expertise. (Section \ref{sec:user-study})
\item \textit{Can the data collected in simulation be effectively \textbf{transferred to real-world} robots?}
We demonstrate that policies trained on data collected through \interface{} transfer zero-shot to real environments with simple Sim2Real techniques. We also highlight the generalizability of \interface{} policies compared to those trained with real-world data. (Section \ref{sec:sim2real})
\end{enumerate}

\subsection{User Study}
\label{sec:user-study}

\begin{table*}[t]
\centering 
\begin{tabular}{@{}lc|ccccc|c@{}}
\toprule
\multirow{2}{*}{} & \multicolumn{1}{l|}{\multirow{2}{*}{}} & \multirow{2}{*}{\begin{tabular}[c]{@{}c@{}}\interface{} \\ \\ \includegraphics[width=0.13\columnwidth]{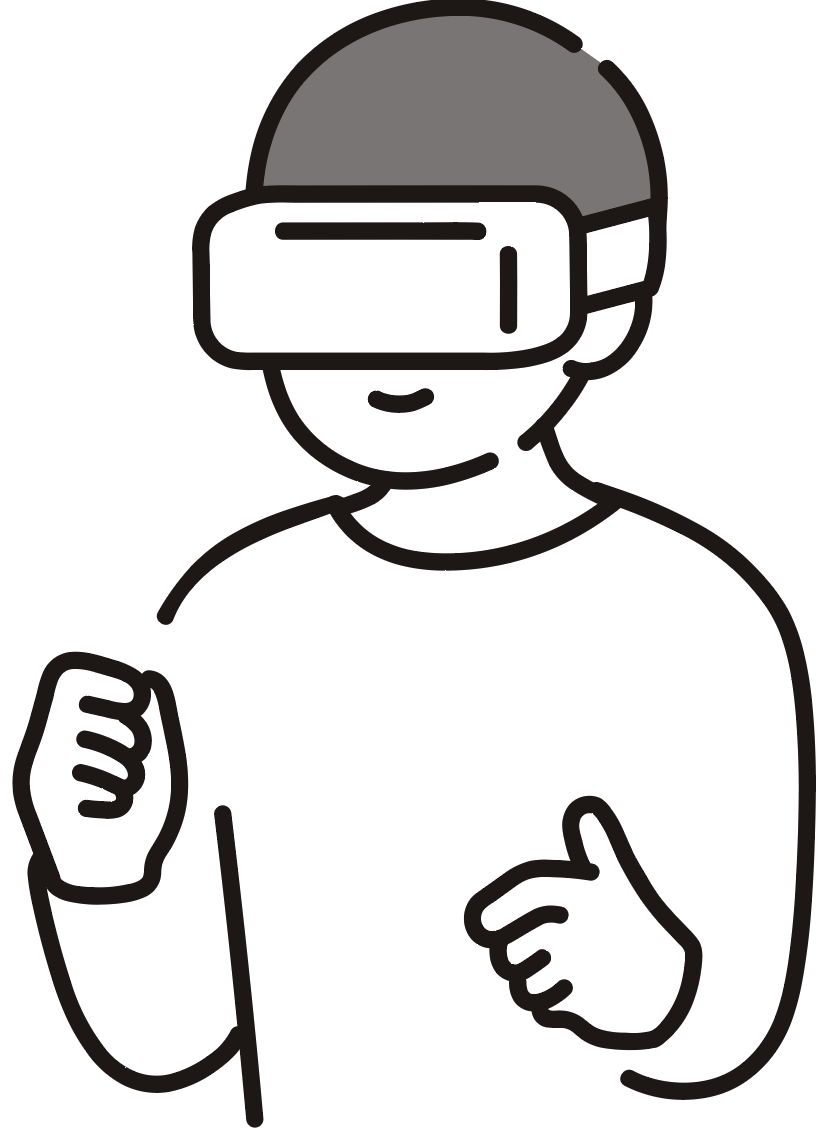}\end{tabular}} &\multicolumn{1}{c}{\begin{tabular}[c]{@{}c@{}}Modulation of\\\textbf{Command Interface}\end{tabular}} & \multicolumn{3}{c|}{\begin{tabular}[c]{@{}c@{}}Modulation of \\\textbf{Visual Feedback Design}\end{tabular}} & \multirow{2}{*}{\begin{tabular}[c]{@{}c@{}} ALOHA \cite{zhao2023learning}\\ \\\includegraphics[width=0.2\columnwidth]{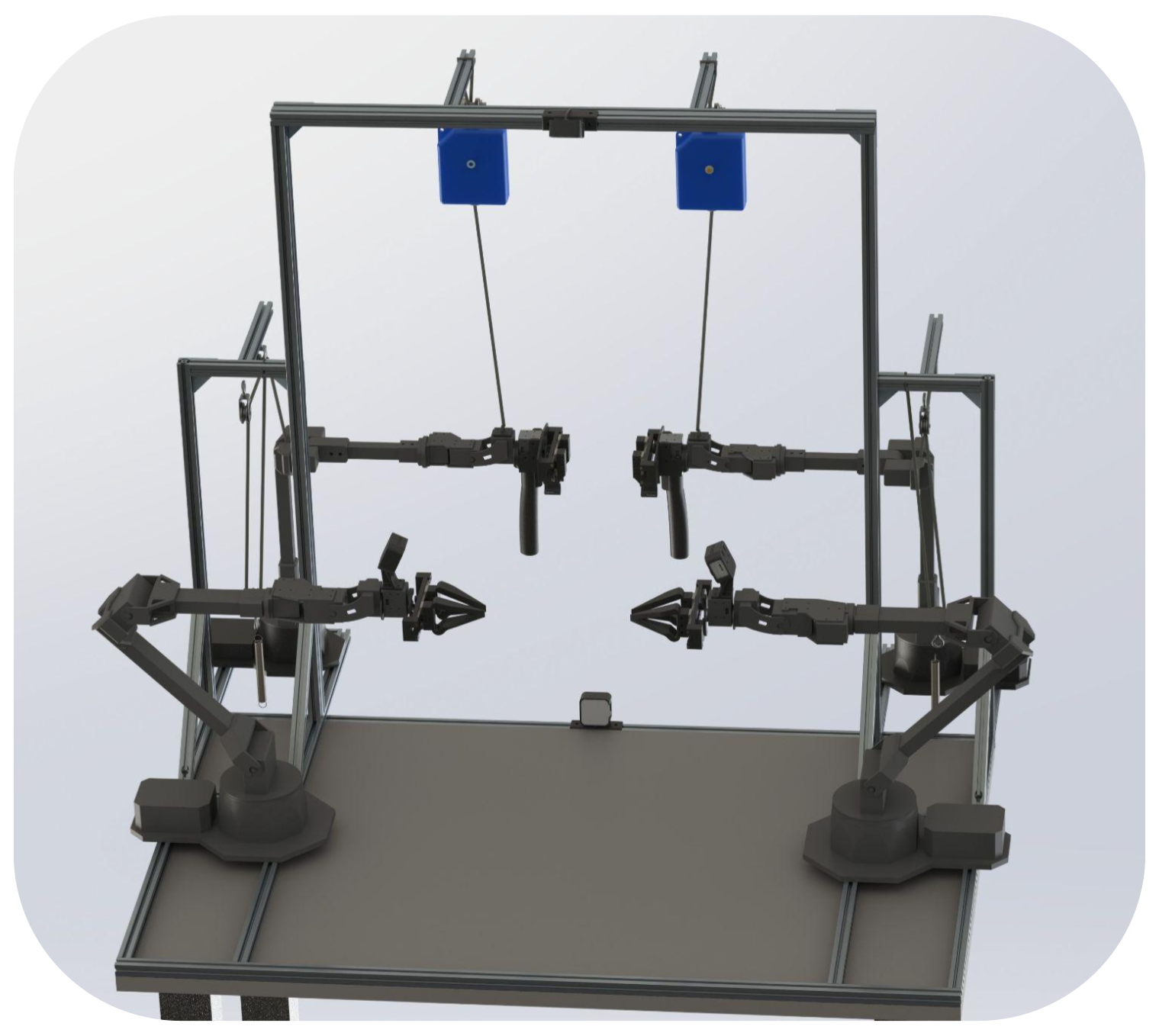}\end{tabular}} \\ \cmidrule(lr){4-7}
  & \multicolumn{1}{l|}{} & & \begin{tabular}[c]{@{}c@{}}Finger\\Tracking \\ $\downarrow$ \\ Kinematic \\Double \end{tabular} & \begin{tabular}[c]{@{}c@{}}Rendering as \\ AR Objects  \\ $\downarrow$ \\ Sim Rendering\\(RGB, Stereo)\end{tabular} & \begin{tabular}[c]{@{}c@{}}Rendering as \\ AR Objects  \\ $\downarrow$ \\Sim Rendering\\(RGB, Mono)\end{tabular} & \begin{tabular}[c]{@{}c@{}}Active\\Viewpoint \\ $\downarrow$ \\ Fixed\\Viewpoint\end{tabular} &  \\ \midrule

\multirow{1}{*} & \begin{tabular}[c]{@{}c@{}}Data \\ Throughput\end{tabular} & \begin{tabular}[c]{@{}c@{}}\textbf{7.8}\\parts / min\end{tabular} & \begin{tabular}[c]{@{}c@{}}6.8\\parts / min\end{tabular} & \begin{tabular}[c]{@{}c@{}}3.6\\parts / min\end{tabular} & \begin{tabular}[c]{@{}c@{}}3.0\\parts / min\end{tabular} & \begin{tabular}[c]{@{}c@{}}2.7\\parts / min\end{tabular} & \begin{tabular}[c]{@{}c@{}}\textbf{3.7}\\parts / min\end{tabular} \\ \bottomrule 
\end{tabular}

\caption{Quantitative comparison between different teleoperation setups for two ViperX arms with parallel-jaw gripper \cite{zhao2023learning}. Users are tasked to organize ten bolts and nuts into two boxes, and \interface{} allowed users to organize 7.77 parts per minute on average, while modulation of both command interface and visual feedback settings dropped the performance significantly. We report percent change in throughput relative to \interface{} averaged across users.} 
\label{table:quant-userstudy-aloha}
\vspace{-1em}
\end{table*}

Through a controlled user study, we analyze the impact of \interface{}'s design decisions on intuitiveness and usability.
Specifically, we compare: (a) the experience of collecting data in real-world versus simulation environments (Sec \ref{sec:user-study-real-vs-sim}), (b) methods of visual perception (Sec \ref{sec:user-study-perception}), and (c) control interfaces (Sec \ref{sec:user-study-control}). A total of nine participants with no prior experience in robotics were recruited.

In varying settings, participants spent 7 minutes collecting as many robot demonstrations as possible. We asked the participants to organize 10 bolts and nuts from a table into boxes. Participants were responsible for resetting the scene both in simulation and real-world environments via reset button or manual effort, respectively.
Participants teleoperated two ViperX arms with parallel-jaw grippers, and kinematically equivalent teacher devices were used as a real-world teleoperation interface \cite{zhao2023learning}.
Quantitative results are presented in Table \ref{table:quant-userstudy-aloha}; further analysis follows.

\subsubsection{Teleoperating in Real-World vs Simulation}
\label{sec:user-study-real-vs-sim} 

\begin{figure}
    \centering
    \includegraphics[width=0.95\columnwidth]{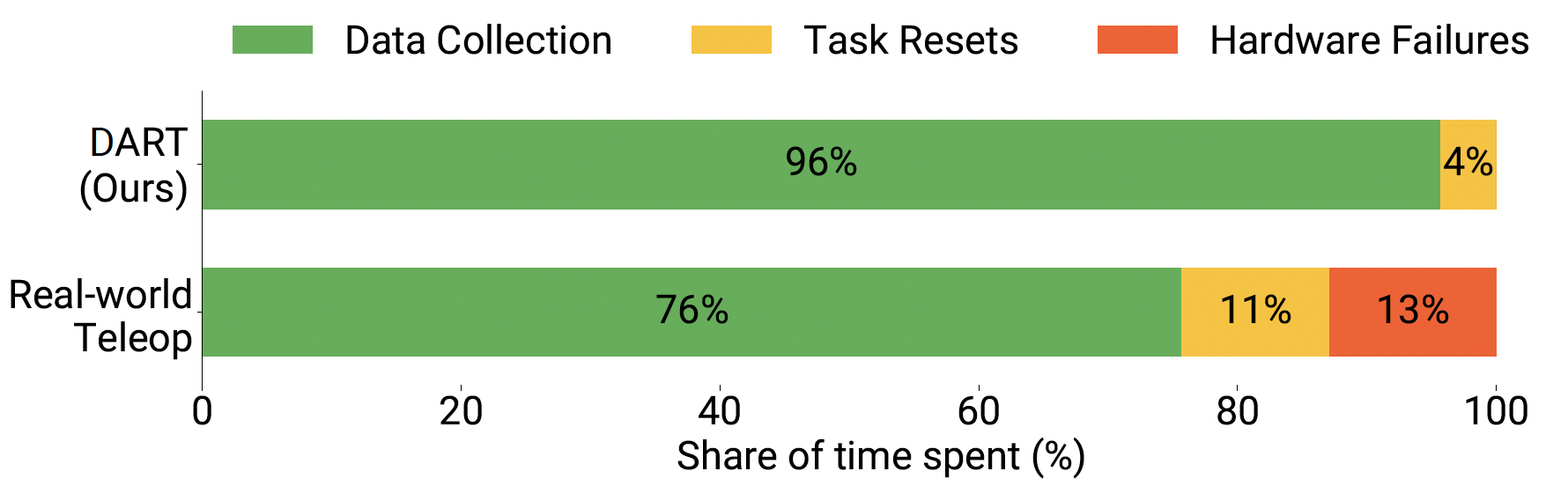}
    \caption{\interface{} allows operators to spend more time on actual data collection, rather than supplementary tasks such as resetting the environment for every task completion or dealing with hardware failures.}
    \label{fig:teleop-time-breakdown}
    \vspace{-1em}

\end{figure}

Our user study comparing \interface{} and real-world teleoperation revealed two key findings. First, a significant portion of time in real-world data collection is spent physically resetting the environment and managing unexpected hardware failures (e.g., performing electrical resets after motor malfunctions) as reported in Fig. \ref{fig:teleop-time-breakdown}. By contrast, most of the time in \interface{} is dedicated to actual data collection.

\begin{figure}
    \centering
    \includegraphics[width=\columnwidth]{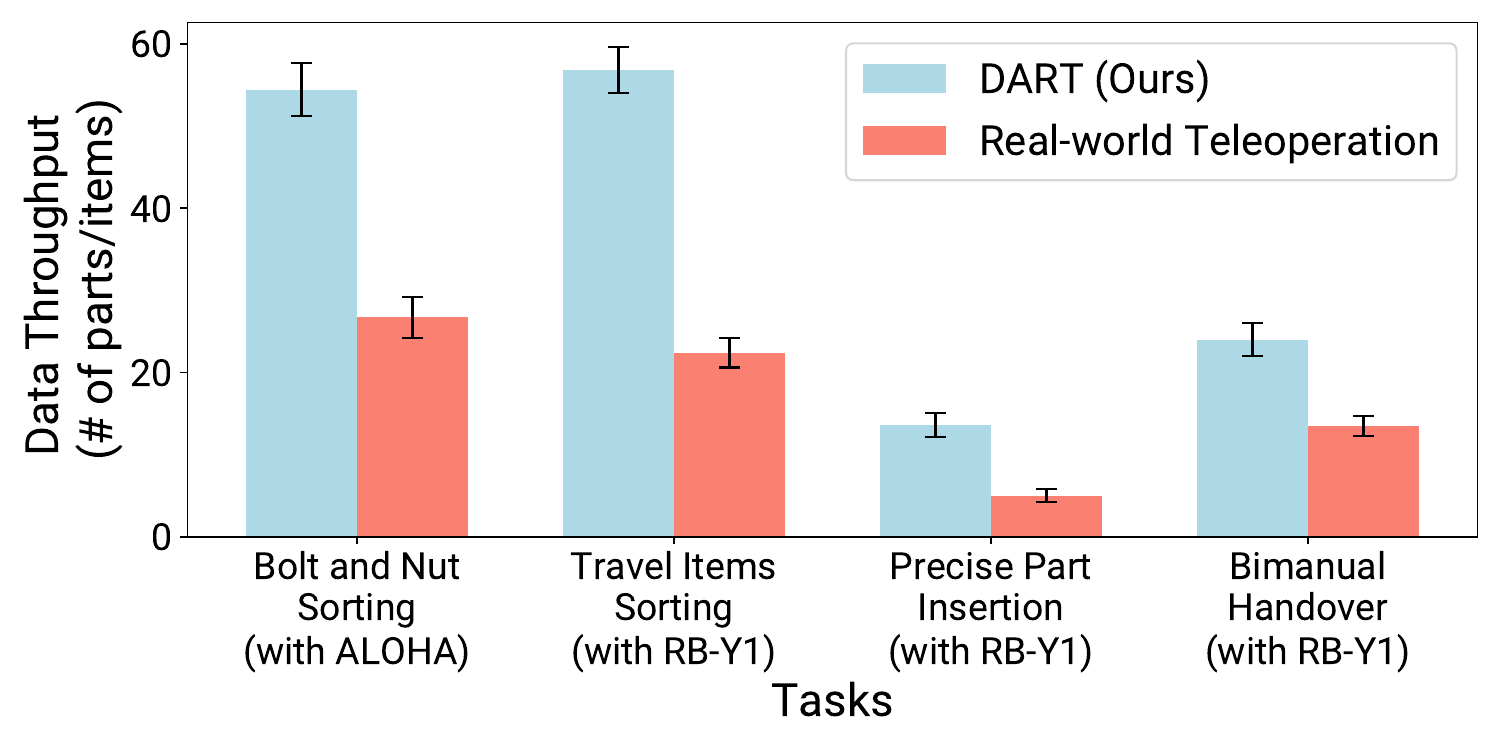}
    \caption{Data throughput comparison between \interface{} and real-world teleoperation systems. For each robot and task, five participants were asked to teleoperate the tasks as many as possible for 7 minutes. For real-world teleoperation, kinematically equivalent teacher device, i.e., kinematic double, was used as a teleoperation interface.}
    \label{fig:user-study-qual}
\end{figure}

Second, even after accounting for reset times and hardware malfunctions, participants in real-world teleoperation showed around \textbf{2$\times$} lower data collection throughput. 
For a comparison experiment with wide range of real-world data collection systems, we used two different robot systems: dual ViperX arms \cite{zhao2023learning} and RB-Y1 from Rainbow Robotics. Both data collection system has kinematic double as its  teleoperation interface. Total 20 participants were asked to perform 4 bimanual tasks ranging from relatively simple object rearrangment task to precise insertion tasks. Figure \ref{fig:user-study-qual} shows the data throughput comparison between \interface{} and two different real-world robot systems on four different tasks.

Many participants attributed this considerable data throughput gap to a) physical fatigue during teleoperation and b) their inability to closely observe local contact interactions, which hindered their ability to perform tasks effectively (Table \ref{fig:user-study-qual}). This particular attribution becomes evident with following ablation studies.

\subsubsection{Effect of Visual Observation on Human Operator's Performance}
\label{sec:user-study-perception}

Our key findings are threefold. First, transmitting images over a network inevitably introduces a tradeoff between latency and decreased visual fidelity, which can negatively impact teleoperation experience. 
All methods transmitting simulation renderings over the network (those with stero and mono rendering) suffered a significant drop in user's data collection throughput compared to \interface{} which transmits only the raw simulation states. 

Second, we find that mono rendering, which limits the ability to properly perceive depth, suffered a performance drop over stereo rendering. Additionally, some participants reported feeling nauseous (Table ~\ref{table:quant-userstudy-aloha}) with stereo rendering -- which uses a fixed interpupillary distance (IPD). By contrast, \interface{} relies on VisionOS's \footnote{Apple's Operating System for AR devices}
native rendering engine, which dynamically adjusts to each user’s IPD  \cite{appleipd}. 

Finally, we found that active perception, where users can explore their surroundings and adjust their viewpoint by moving their heads, is critical. Teleoperation without active perception reduces the data collection rate by $21.7\%$.

\begin{figure}
    \centering
    \includegraphics[width=\columnwidth]{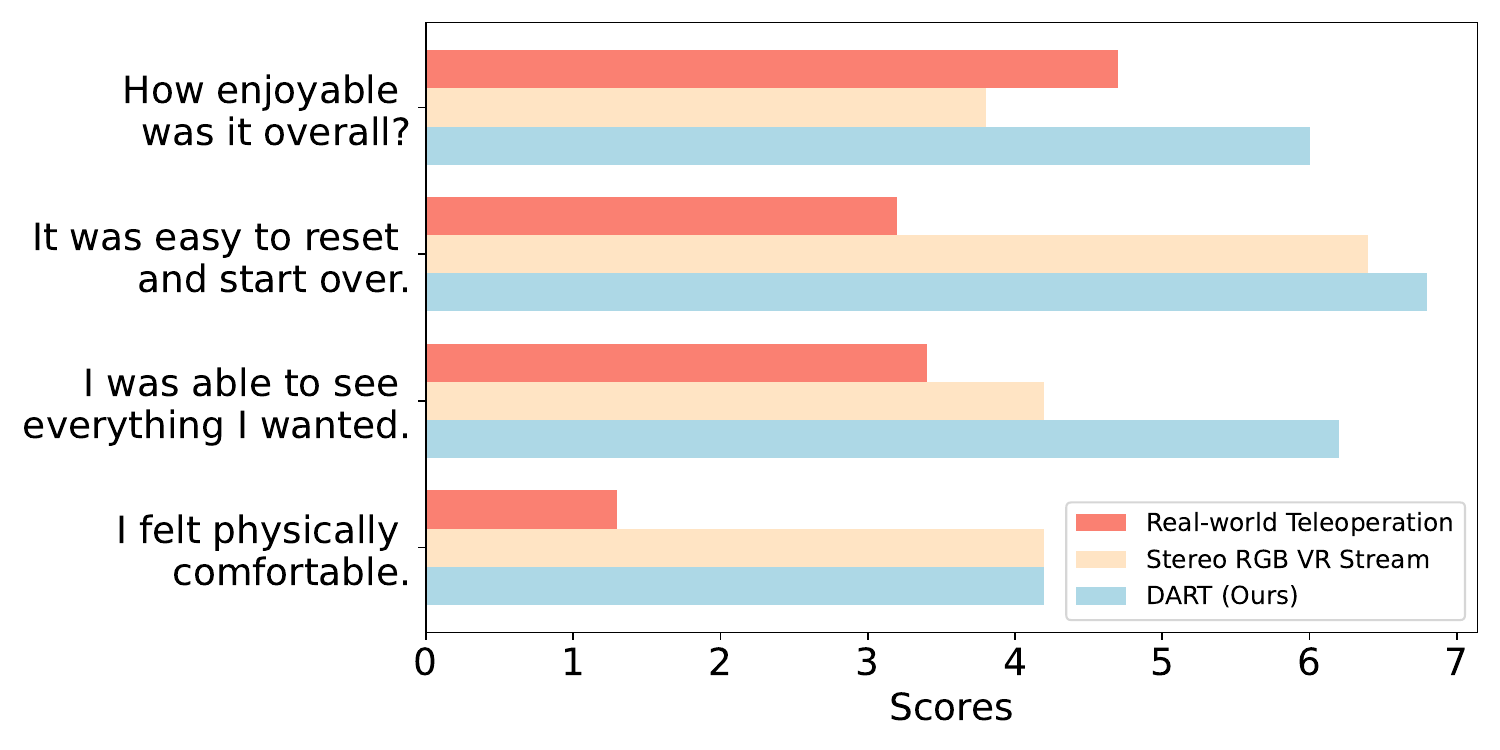}
    \caption{Qualitative comparison between different teleoperation interfaces amongst user study participants. Participants reported that \interface{} is enjoyable, physiaclly less fatiguing and allows better visual observation during teleoperation.}
\end{figure}

\subsubsection{Control Method}
\label{sec:user-study-control}

We compared two methods for operating robots in simulation: a) a kinematically equivalent teacher device and b) inverse kinematics (IK) using hand tracking keypoints as targets. Our findings indicate that the kinematic double did not significantly improve task success rate over its IK equivalent. While the kinematic double provides more direct control over the robot’s joints, users reported that the intuitive hand tracking offered by \interface{} was sufficient, or even better, due to reduced weight and strain on the operator (Table \ref{table:quant-userstudy-aloha}).

\subsection{Sim2Real and Generalizability}
\label{sec:sim2real}

\begin{figure}[t]
    \centering
    \begin{minipage}{0.45\columnwidth}
        \centering
        \includegraphics[width=\linewidth]{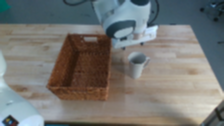}
        \subcaption{Nominal Lab Setting}
        \label{fig:sub1}
    \end{minipage}
    \begin{minipage}{0.45\columnwidth}
        \centering
        \includegraphics[width=\linewidth]{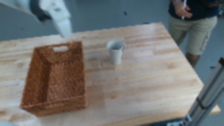}
        \subcaption{Camera Pose Change}
        \label{fig:sub2}
    \end{minipage}
    
    \vspace{0.1cm}
    
    \begin{minipage}{0.45\columnwidth}
        \centering
        \includegraphics[width=\linewidth]{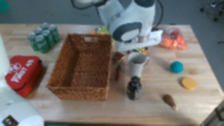}
        \subcaption{Unseen Distractions}
        \label{fig:sub3}
    \end{minipage}
    \begin{minipage}{0.45\columnwidth}
        \centering
        \includegraphics[width=\linewidth]{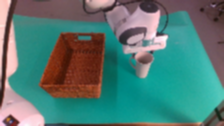}
        \subcaption{Green background}
        \label{fig:sub4}
    \end{minipage}
    
    \vspace{0.1cm}
    
    \begin{minipage}{0.45\columnwidth}
        \centering
        \includegraphics[width=\linewidth]{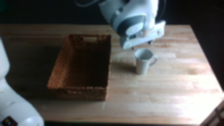}
        \subcaption{Lighting Change}
        \label{fig:sub5}
    \end{minipage}
    \begin{minipage}{0.45\columnwidth}
        \centering
        \includegraphics[width=\linewidth]{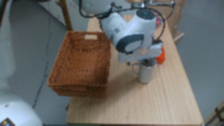}
        \subcaption{Location Change}
        \label{fig:sub6}
    \end{minipage}
    
    \caption{Six different settings to evaluate the robustness of our RGB vision-based policy trained with data collected through \interface{}.}
    \label{fig:different-setups}
    \vspace{-2em}
\end{figure}

Both \interface{} and real-world data collection offer distinct advantages for real-world policy training.
With \interface{}, roboticists benefit from significantly higher data throughput with reduced physical and cognitive demands, as demonstrated by our user study (Sec. \ref{sec:user-study}). 
One minor downside of using \interface{} is the need to import scenes into the simulation environment. Fortunately, with modern advances in computer vision \cite{daneshmand20183d, hampali2023hand}, scanning 3D objects from the real world has become incredibly efficient. The bigger challenge, however, lies in bridging the potentially large Sim2Real gap. Given these trade-offs, how does one weigh the benefits of faster data collection against the challenge of real-world deployment?

Our experimental results suggest that collecting data in simulation offers \textbf{more advantages than drawbacks} when paired with a proper Sim2Real pipeline. In particular, we demonstrate the unique robustness of Sim2Real-transferred policies, enabled by diverse data augmentation techniques only available in simulation environments.

\begin{figure}[t]
    \centering
    \begin{subfigure}[b]{0.9\columnwidth}
        \centering
        \includegraphics[width=\textwidth]{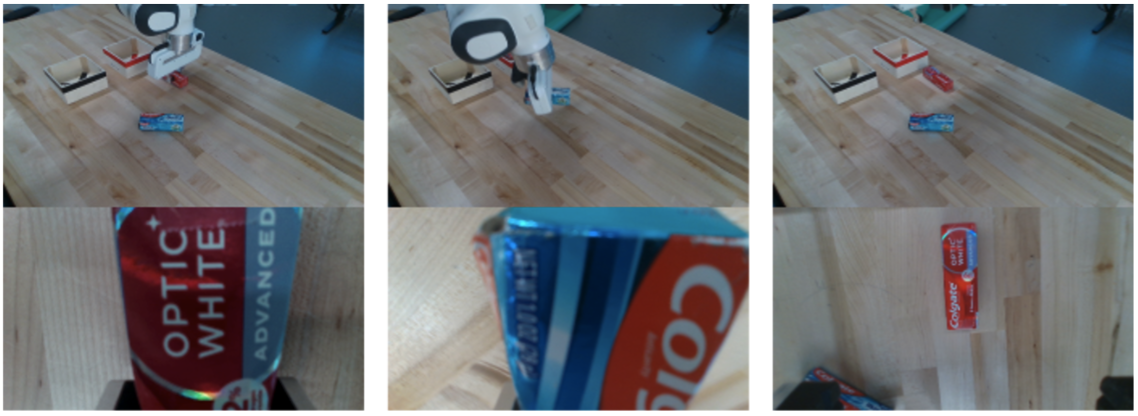}
        \caption{Real-world Images}
        \label{fig:subfig1}
    \end{subfigure}
    \vspace{5pt}
    \begin{subfigure}[b]{0.9\columnwidth}
        \centering
        \includegraphics[width=\textwidth]{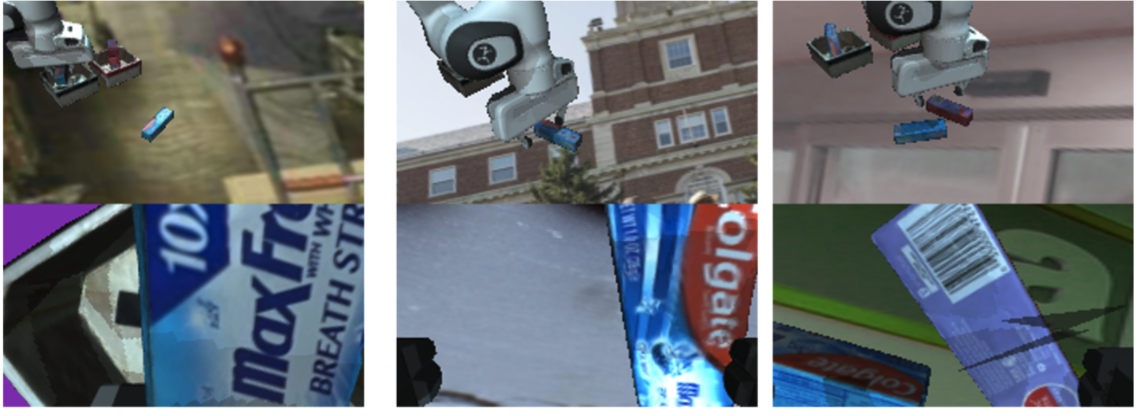}
        \caption{Simulation Renderings with Background Augmentations}
        \label{fig:subfig2}
    \end{subfigure}
    \caption{
    Visual comparison between training images for Real and \interface{} policies. Simulation allows augmentation out-of-the-box, which results in zero-shot Sim2Real and robustness.
    }
    \label{fig:sim2real}
\end{figure}

Specifically, we compare two types of RGB vision policies: (a) a policy trained on real-world data, and (b) a policy trained on simulation data collected through \interface{}. Both policies are trained on two tasks with 50 minutes of operator effort. Both policies also use a standard ACT \cite{zhao2023learning} implementation at 20Hz. Real-world datasets are augmented with Gaussian blur and color-jitter. \interface{} datasets were additionally augmented by randomizing the camera extrinsic and intrinsics, replacing the background with random textures and images from \cite{dtd, openimagesv4, indoorscenes}, and randomizing the lighting setting in simulation (Figure \ref{fig:sim2real}).

Inspired by \cite{xie2024decomposing}, we evaluated policies in six diverse environments in the real world illustrated in Fig. \ref{fig:different-setups}. We found that our \interface{} policies not only demonstrate zero-shot Sim2Real in the nominal setting but also significantly outperform the Real policy in many of the modified settings (Table \ref{tab:sim2real_results}). Our results highlight the benefit of scaling up simulation data versus real-world data: a single demo in simulation, which can be aggressively augmented, is more valuable for learning than that collected in real world.

\begin{table}
\centering
\footnotesize
\begin{tabular}{@{}l@{\hspace{0.5em}}l|cc|cc@{}}
\toprule
& ~Task & \multicolumn{2}{c|}{Pick Mug in Basket} & \multicolumn{2}{c}{Sorting Small Items} \\ 
\cmidrule(l){3-6}
\multicolumn{2}{@{}c|}{Trained on data from} & Real-world & \interface{} & Real-world & \interface{} \\
\midrule
\multicolumn{2}{@{}l|}{~Lab Space} & 65\% & \textbf{80\%} & 45\% &\textbf{60\%} \\
& \textit{with}: &&&& \\
& ~~~Lighting Changes & 45\% & \textbf{45\%} & 25\% & \textbf{65\%} \\
& ~~~Background Changes & 10\% & \textbf{60\%} & 5\% & \textbf{40\%} \\
& ~~~Cam. Pose Changes & 5\% & \textbf{35\%} & 0\% & \textbf{40\%} \\
& ~~~Unseen Distractions & 0\% & \textbf{70\%} & 0\% & \textbf{45\%} \\
\midrule
\multicolumn{2}{@{}l|}{Communal Kitchen} & 0\% & \textbf{50\%} & 0\% & \textbf{35\%} \\
\bottomrule
\end{tabular}
\caption{Success rates for policies trained with 50 minutes of data collection effort in the real-world v/s \interface{}. The results highlight the robustness of policies trained with simulation data, enabled by diverse data augmentation strategies.}
\label{tab:sim2real_results}
\vspace{-2em}
\end{table}

%% file: dexhub.tex
\section{DexHub: Central Data Hub \\for Robot Learning on the Cloud}
\label{sec:dexhub}

\subsection{Purpose and Vision}

To serve as a central data hub for logging \textit{every} demonstration collected through \interface{}, we developed \textbf{DexHub}, a cloud-hosted data repository where anyone can sign in and retrieve datasets collected by themselves and others.

In fact, to further enhance its role as an organically growing data hub, DexHub also provides an API that enables users to log all robot interaction with ease, regardless of whether they use \interface{} or other setups. Leveraging a cloud database, user authentication system, and secure data logging, the API allows seamless integration for individuals and institutions alike to contribute and access data. The user authentication system ensures that every data contribution is properly attributed to the individual who made it, offering potential for future reward mechanisms based on contributions.

\subsection{API for End-Users}

DexHub's token-protected API supports multiple key functionalities ranging from downstream (downloading from the cloud) and upstream (uploading to the cloud) operations.

\subsubsection{Downstream API} 

Users can retrieve the data they have personally collected through \interface{} by simply hitting $\texttt{/get-my-data}$ with an API key retrieved from our \href{https://dexhub.ai/}{website}. This endpoint returns a list of downloadable links for every log file that users have uploaded to the cloud. The API also allows users to access the \textit{global} dataset which includes robot data collected and contributed by other users. Global dataset will be made available upon curation for public use.

\subsubsection{Upstream API} 

We provide an easy-to-use upstream API allowing users to contribute to DexHub without an AR device. A simple addition of $\texttt{dexhub.log(obs, act)}$ to any Python-based robot execution script will automatically log and upload robot interactions to DexHub. 
All upstream contributions will be logged in the system and properly attributed to the individual who contributed. To retrieve the API keys and learn more about the detailed usage instructions, visit \url{https://dexhub.ai}.

%% file: conclusion.tex
\section{Discussions}

In this paper, we present \interface{}, \underline{D}exterous \underline{A}ugmented \underline{R}eality \underline{T}eleoperation system, enabling inutitive, low-latency teleoperation in cloud-hosted simulation.
We believe that \interface{}'s intuitive teleoperation interface combined with \platform{}'s versatile data logging features will pave the path towards an internet-scale, ever-growing robot learning dataset.

However, \interface{} has a few limitations -- mostly stemming from the limitation of physics simulation itself. Tasks that cannot be simulated by current physics engines, e.g., chopping an onion, cannot be demonstrated in \interface{}. Deformable objects, although not impossible, are still hard to simulate.

The rapid advancements in physics engines and simulation technologies make us confident that these barriers will diminish over time. It is also important to note that we are not suggesting simulation as the sole path forward. Real-world datasets remain invaluable, and \interface{} is designed to complement rather than replace them. By supporting both simulated and real-world data collection through \platform{}, we aim to strike a balance that leverages the strengths of each approach.